\documentclass{article}

\PassOptionsToPackage{numbers, compress}{natbib}

\usepackage{arxiv}
\usepackage{natbib}

\usepackage[utf8]{inputenc} 
\usepackage[T1]{fontenc}    
\usepackage{hyperref}       
\usepackage{url}            
\usepackage{booktabs}       
\usepackage{amsfonts}       
\usepackage{nicefrac}       
\usepackage{microtype}      
\usepackage{xcolor}         

\usepackage{bm}
\usepackage{calc}
\usepackage{amsmath}
\usepackage{wrapfig}
\usepackage{enumerate}
\usepackage{enumitem}
\usepackage{multirow}
\usepackage{graphicx}
\usepackage{subcaption}
\usepackage{algorithm}
\usepackage{algpseudocode}

\hypersetup{
  colorlinks=true,
  citecolor=[HTML]{F59E32},
  linkcolor=[HTML]{9fd9a7},
  urlcolor=magenta!70!black 
}

\newcommand{\gray}[1]{\textcolor{gray}{#1}}

\title{Dynamic Chunking for Diffusion Language Models}

\author{%
  \textbf{Yichen Zhu} \\ CSE, HKUST \\ \texttt{yc\_zhu@zju.edu.cn}
  \And
  \textbf{Xiaoming Shi} $^{\dagger\ddagger}$ \\ Xiaohongshu Inc. \\ \texttt{sxm728@hotmail.com}
  \And
  \textbf{Peng Zhao} \\ Alibaba group \\ \texttt{zhuyun.zp@alibaba-inc.com}
  \AND
  \textbf{Weiyu Chen} \\ CityUHK \\ \texttt{weiyu.chen@cityu.edu.hk}
  \And
  \textbf{Debing Zhang} \\ Xiaohongshu Inc. \\ \texttt{dengyang@xiaohongshu.com}
  \And
  \textbf{James Kwok} \\ CSE, HKUST \\ \texttt{jamesk@cse.ust.hk}
}

\begin{document}

\maketitle

\begin{abstract}
    Block discrete diffusion language models factorize a sequence autoregressively over fixed-size positional blocks, decoupling within-block parallel denoising from across-block conditioning. We argue that this rigid partition wastes structure already present in the sequence: blocks defined by position rather than by content separate semantically coherent tokens and group unrelated ones together. We introduce the \textbf{D}ynamic \textbf{C}hunking \textbf{D}iffusion \textbf{M}odel (DCDM), which replaces positional blocks with content-defined semantic chunks. At its core is Chunking Attention, a differentiable layer that routes tokens into $K$ clusters parameterized by learnable subspaces and shaped end-to-end by the diffusion objective. The resulting cluster assignments induce a chunk-causal attention mask under which a discrete diffusion denoiser factorizes the sequence likelihood autoregressively over semantic chunks, strictly generalizing block discrete diffusion. On downstream benchmarks at parameter scales up to 1.5B, DCDM consistently improves over both unstructured and positional-block diffusion baselines, with the advantage stable across scales and visible early in training.
\end{abstract}

\section{Introduction}
\label{sec:introduction}

\footnotetext[2]{$\dagger$ Project leader.}
\footnotetext[3]{$\ddagger$ Corresponding author: sxm728@hotmail.com}

Diffusion large language models (dLLMs) have recently emerged as a competitive paradigm for text generation, due to their ability to decode multiple tokens in parallel.
Open-source masked diffusion language models (MDLMs)~\cite{mdlm} such as LLaDA~\cite{llada} and Dream~\cite{dream} have achieved performance comparable to autoregressive models at similar scales,
while proprietary models such as 
Gemini Diffusion~\cite{gemini_diffusion} and Mercury~\cite{mercury} demonstrate substantially higher generation throughput. A key ingredient behind this progress is block diffusion~\cite{bdlm}, which has become the dominant design for scalable diffusion-based language modeling.

Block diffusion combines the strengths of autoregressive and diffusion models. It factorizes a sequence autoregressively over blocks, preserving causal conditioning across groups of tokens, while denoising all tokens inside each block bidirectionally and in parallel. This design provides a practical compromise between the quality of autoregressive modeling and the efficiency of parallel diffusion sampling. However, existing block diffusion language models (BDLMs)~\citep{bdlm} define blocks by a fixed positional rule: a sequence is partitioned into contiguous segments of equal length. This choice imposes a strong structural prior that is independent of the content of the sequence.

We argue that fixed positional blocks are a limiting abstraction for language. The dependencies that determine a token are often not aligned with local contiguity: an entity may govern distant mentions, a mathematical derivation may depend on earlier premises, and a code token may be constrained by scope or syntax several lines away. Positional partitioning can therefore separate tokens that should be denoised jointly, while placing weakly related neighboring tokens in the same diffusion process.
In this case, the model inherits the block-autoregressive factorization but applies it at a granularity not matched to the semantic structure of the sequence.

To address this mismatch, we propose the \emph{Dynamic Chunking Diffusion Model} (DCDM), which replaces fixed positional blocks with learned semantic chunks. Rather than segmenting a sequence by position, DCDM clusters tokens according to representations produced inside the model. The resulting chunks may be non-contiguous, variable in size, and sequence-dependent. They serve the same role as blocks in block diffusion: tokens within a chunk are denoised bidirectionally in parallel, while chunks are ordered autoregressively through a chunk-causal mask. Thus, DCDM preserves the computational structure of block diffusion while making the unit of parallel denoising content-adaptive.

The core component of DCDM is \emph{Chunking Attention}, a differentiable routing layer that assigns tokens to one of $K$ chunks. Direct point-centroid clustering is unstable in high-dimensional language-model representations, as a few clusters can dominate early and starve the remaining ones of gradient signal. We instead represent each chunk by a learnable low-dimensional subspace and route tokens according to subspace alignment~\cite{subspace_clustering}.
A soft attention path places the chunking geometry directly on the gradient path of the diffusion objective, while the induced hard assignments define a semantic chunk-causal attention mask. This construction generalizes positional block diffusion, which is recovered when the learned chunks coincide with fixed contiguous blocks.

The contributions of this paper are:
\begin{itemize}[leftmargin=*,itemsep=2pt,topsep=2pt]
    \item We introduce {Chunking Attention}, a subspace-based differentiable routing mechanism that induces semantic chunk-causal masks for diffusion language modeling.
    \item We develop DCDM, a diffusion language model that strictly generalizes BDLM and trains the chunking mechanism and denoiser end-to-end under the diffusion objective.
    \item We provide extensive empirical evidence that semantic chunking outperforms its positional counterpart on downstream tasks across general reasoning, mathematics, and code generation at both 0.5B and 1.5B scales, validating that block diffusion benefits substantially from content-adaptive granularity.
\end{itemize}


\section{Related Work}

\paragraph{Diffusion Large Language Models.}
The landscape of language generation has long been dominated by autoregressive models~\cite{gpt, gpt2, gpt4, qwen, llama, deepseek, gemini}. 
While celebrated for their high-quality outputs, these models are fundamentally constrained by a sequential, token-by-token decoding process~\cite{ar_diff_model, aoar}. 
To alleviate these latency bottlenecks, 
dLLMs, a class of diffusion-based frameworks specifically designed for the discrete data domain, 
have emerged as a compelling alternative. 
By incorporating an absorbing state, e.g., \texttt{[MASK]}, to represent noise, \citet{d3pm} laid the foundation for masked diffusion modeling. 
This framework has been subsequently extended by a series of recent works~\cite{mdlm, md4, llada, dream, diffugpt}.
Notably, MDLM~\cite{mdlm} is among the most widely adopted, offering a simple yet highly efficient training objective. 
The LLaDA~\cite{llada} series scales diffusion language models beyond 8 billion parameters, demonstrating performance comparable to, if not exceeding, that of autoregressive models of equivalent scale.

\paragraph{Autoregression-diffusion Hybrid Language Models.}
Recent works have explored integrating the computational efficiency of autoregressive models into diffusion-based frameworks, 
particularly for complex tasks such as video synthesis. 
A representative approach, 
BDLM~\cite{bdlm}, models semantic dependencies across blocks autoregressively while performing the denoising process independently within each block. 
Fast-dLLMs~\cite{fast_dllm} employ techniques such as block-wise prefix caching to achieve generation efficiency that substantially surpasses that of AR models, without compromising generation quality.
Another line of work attempts to relax the rigidity of fixed positional blocks at \emph{inference time}. 
AdaBlock-dLLM~\cite{adablock} adaptively adjusts block boundaries during sampling using local denoising-confidence signals.


\section{Preliminary}
\label{sec:preliminary}


\subsection{Masked Diffusion Models}

Masked Diffusion Language Models (MDLMs)~\cite{mdlm} are a class of discrete diffusion models in which the absorbing distribution $\bm{\pi}$ of the forward process is the point mass on a special mask token $\mathbf{m}$.
Let $\mathbf{x} \in \mathcal{V}^L$ denote a clean sequence of length $L$ drawn from the data distribution, and let $\mathbf{z}_t$ denote its corrupted latent variable at timestep $t \in [0, 1]$.
The forward process operates over continuous time $t \in [0, 1]$ and replaces each token independently by $\mathbf{m}$ with probability $1 - \alpha_t$, where $\alpha_t$ is a predefined noise schedule strictly decreasing from $\alpha_0 = 1$ (clean data) to $\alpha_1 = 0$ (fully masked). 
The reverse process is parameterized by a denoiser $\mathbf{x}_\theta(\mathbf{z}_t, t)$ trained to predict clean data from the masked state, and its evidence lower bound (ELBO) simplifies to the per-sample weighted cross-entropy loss
\begin{equation}
\label{eq:mdlm_loss}
    \mathcal{L}(\mathbf{x}, \theta)
    \;=\; \mathbb{E}_{q(\mathbf{z}_t \mid \mathbf{x})}
    \int_{0}^{1}
    \frac{\alpha_t'}{1 - \alpha_t}
    \sum_{\ell:\, \mathbf{z}_t^{\ell} = \mathbf{m}}
    \log \bigl\langle
        \mathbf{x}_{\theta,\ell}(\mathbf{z}_t^{1:L}, t),\;
        \mathbf{x}_{\ell}
    \bigr\rangle \, \mathrm{d}t,
\end{equation}
where $\alpha_t' = \mathrm{d}\alpha_t/\mathrm{d}t$ and $\mathbf{x}_{\theta,\ell}$ is the predicted categorical distribution at position $\ell$.

\subsection{Block Diffusion Models}

Block Diffusion Language Models (BDLMs)~\cite{bdlm} combine autoregressive
and diffusion modelling by partitioning a sequence of $L$ tokens into $K$
contiguous blocks of fixed length $B$ (with $L = K \cdot B$), performing
discrete diffusion within each block while maintaining autoregressive
dependencies across blocks. The likelihood factorizes autoregressively over
blocks,
\begin{equation}
\label{eq:bdlm_likelihood}
    \log p_\theta(\mathbf{x})
    \;=\; \sum_{b=1}^{K}
    \log p_\theta\!\left(\mathbf{x}^{b} \,\big|\, \mathbf{x}^{<b}\right),
\end{equation}
where $\mathbf{x}^{b}$ denotes the $b$-th block of $B$ tokens and
$\mathbf{x}^{<b}$ denotes all preceding blocks. Each conditional
$p_\theta(\mathbf{x}^{b} \mid \mathbf{x}^{<b})$ is modeled by a discrete
diffusion process over the $B$ tokens of that block. Applying the masked
diffusion ELBO to each block yields the per-sample training bound
\begin{equation}
\label{eq:bdlm_elbo}
    \log p_\theta(\mathbf{x})
    \;\geq\; \mathcal{L}_{\text{BD}}(\mathbf{x}, \theta)
    \;:=\; \sum_{b=1}^{K}
    \mathcal{L}\!\left(
        \mathbf{x}^{b}, \mathbf{x}^{<b}, \theta
    \right),
\end{equation}
By tuning the block size $B$, BDLMs interpolate between pure diffusion
($B = L$, $K = 1$) and autoregressive models ($B = 1$, $K = L$). This
design supports flexible-length generation, KV caching across blocks, and
parallel sampling within each block, while achieving state-of-the-art
perplexity among discrete diffusion models.


\begin{figure}[t]
  \centering
  \includegraphics[width=\textwidth]{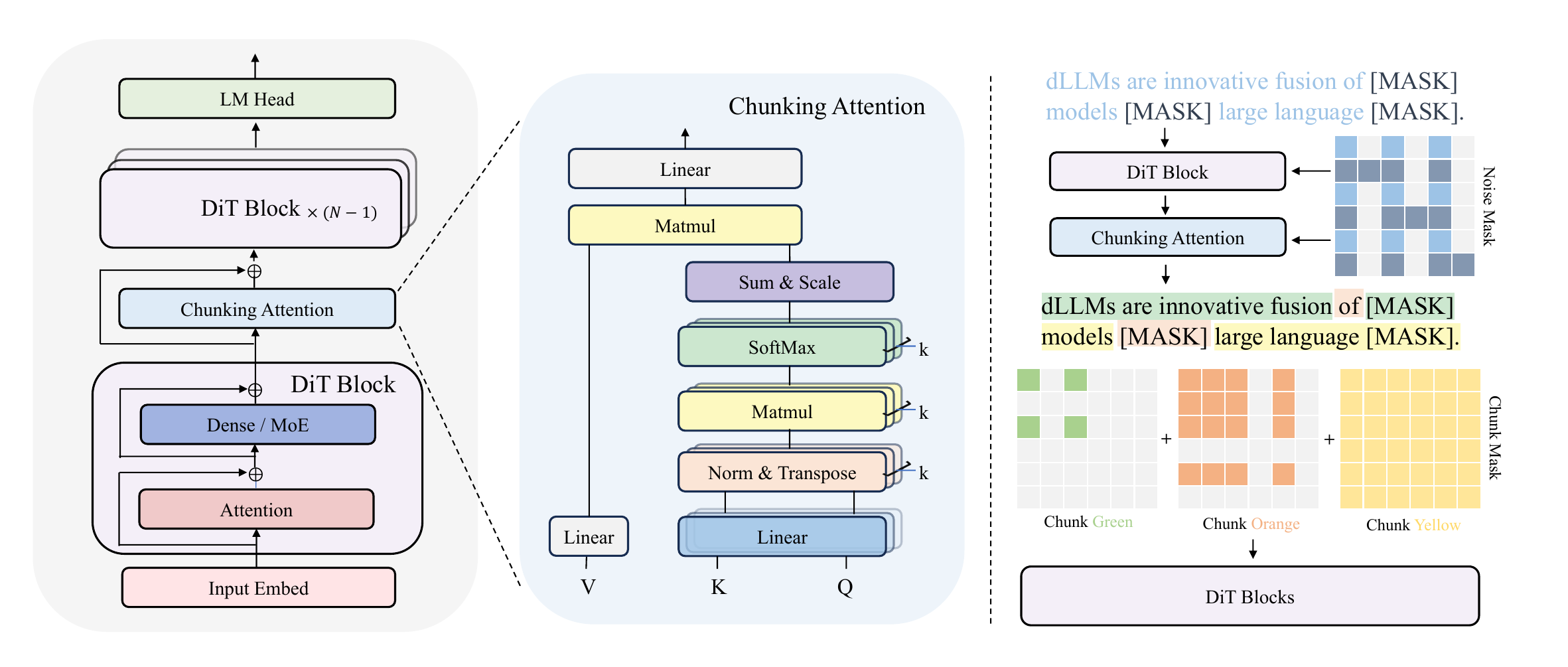}
  \caption{
    Overview of DCDM. \textbf{Left:} The denoiser stacks $N{-}1$ DiT blocks on top of a single Chunking Attention layer that produces the content-defined partition consumed by all downstream blocks.
    \textbf{Right:} Operationally, the chunking attention takes the noisy input together with a noise mask and emits a per-token cluster assignment (color-coded as Green, Orange, Yellow), which induces the chunk-causal attention mask used by every subsequent layer of the denoiser.
  }
  \label{fig:overview}
\end{figure}

\section{Methodology}
\label{sec:methodology}

In this section, 
we begin by introducing Chunking Attention in Section~\ref{sec:cha}. 
Building upon this foundational module, Section~\ref{sec:dcdm} presents the Dynamic Chunking Diffusion Model (DCDM) and its training paradigm. 
Finally, Section~\ref{sec:load_balancing} describes the load-balancing mechanisms that keep the learned partition well-conditioned throughout training.

\subsection{Chunking Attention}
\label{sec:cha}

We add a new chunking attention layer that performs token-level clustering of hidden states $\mathbf{H} \in \mathbb{R}^{L \times d}$ into $K$ groups,
replacing the positional partition of BDLMs~\citep{bdlm} with one produced inside the model itself. 
Existing block diffusion partitions a sequence into fixed-size positional blocks and applies autoregressive conditioning across them, 
but the positional partition is rigid and cannot adapt to the semantic content of the sequence, 
so semantically related tokens routinely end up split across block boundaries while unrelated tokens are denoised in parallel.

A natural realization of such a partition is to perform clustering inside an attention layer itself: learn a single point centroid $\bm{c}_k \in \mathbb{R}^{d}$ per cluster and route each token by its inner-product similarity to $\bm{c}_k$, an idea explored in prior work on attention-based clustering~\citep{abc}. 
While this recipe has been demonstrated on low-dimensional problems,
we find it unstable in the high-dimensional embedding spaces of modern language models (see Appendix~\ref{app:point-vs-subspace}).
We therefore promote each cluster from a single point to a low-dimensional subspace, adopting the design principle:
\textit{
    Each cluster is characterized by a low-dimensional subspace of
    $\mathbb{R}^{d}$, and tokens are assigned to clusters according to
    their alignment with these subspaces.
}

We parameterize the layer with $K$ learnable matrices $\{\bm{\mu}_k\}_{k=1}^{K}$,
where each $\bm{\mu}_k \in \mathbb{R}^{d \times h}$ serves as the basis of one cluster.
Algebraically, $\bm{\mu}_k$ acts as a projection $\mathbb{R}^{d} \mapsto \mathbb{R}^{h}$ that maps a token $\bm{x}_\ell$ to
\begin{equation}
    \bm{p}_{k,\ell} \;=\; \bm{\mu}_k^{\top}\,\bm{x}_{\ell} \;\in\; \mathbb{R}^{h}.
\end{equation}
Geometrically, the column span
$\mathcal{S}_k := \mathrm{col}(\bm{\mu}_k) \subset \mathbb{R}^{d}$ is an $h$-dimensional subspace associated with the $k$-th cluster, 
in the spirit of subspace clustering~\citep{subspace_clustering}, a line of
work with a long history of success in prior literature~\citep{dscn, aasc}.
Under this view,
$\bm{p}_{k,\ell}$ records the coordinates of $\bm{x}_\ell$ along the $h$ axes spanning $\mathcal{S}_k$, and $\|\bm{p}_{k,\ell}\|$ measures the alignment of $\bm{x}_\ell$ with $\mathcal{S}_k$.
Tokens are free to vary along these $h$ directions without incurring a distance penalty, so intra-cluster variability is absorbed by the geometry of the subspace itself.

To isolate intra-cluster interactions among tokens, the module computes,
for each cluste $k$, a pairwise affinity matrix $\mathbf{A}_k \in \mathbb{R}^{L \times L}$ whose entries are inner products of the projected tokens:
\begin{equation}
    [\mathbf{A}_k]_{\ell,m}
    \;=\; \frac{1}{\sqrt{h}}\, \bm{p}_{k,\ell}^{\top}\, \bm{p}_{k,m}
    \;=\; \frac{1}{\sqrt{h}}\, \bm{x}_\ell^{\top}\, \bm{\mu}_k\, \bm{\mu}_k^{\top}\, \bm{x}_m.
\label{eq:bilinear-gate}
\end{equation}
The factor $1/\sqrt{h}$ is the standard dot-product scaling~\citep{attention}, applied at the projected dimension $h$ rather than $d$ since the inner product is taken in $\mathbb{R}^{h}$.
Sharing $\bm{\mu}_k$ across the query and key sides turns the affinity into a bilinear gate:
$[\mathbf{A}_k]_{\ell,m}$ activates only when $\bm{x}_\ell$ and 
$\bm{x}_m$ are simultaneously aligned with $\mathcal{S}_k$, 
so pairs straddling different clusters are suppressed.

The $K$ per-cluster softmax-ed attention matrices
$\{\mathbf{T}_k = \mathrm{softmax}(\mathbf{A}_k)\}$ are aggregated and applied
to the hidden states to produce the module output $\mathbf{Y}$:
\begin{equation}
    \mathbf{Y} \;=\; \mathbf{W}_O \left(
        \frac{1}{\sqrt{K}} \sum_{k=1}^{K} \mathbf{T}_k
    \right) \mathbf{W}_V\, \mathbf{H},
\label{eq:aggregation}
\end{equation}
where $\mathbf{W}_V, \mathbf{W}_O \in \mathbb{R}^{d \times d}$ are learnable value/output projections independent of the clustering geometry, 
and $1/\sqrt{K}$ stabilizes the variance of the combined operator. 
The purpose of this aggregation is not merely to mix tokens, but to fold the clustering operation into the main computational path of the model: by routing the hidden states through $\sum_{k} \mathbf{T}_k$,
the learnable matrices $\{\bm{\mu}_k\}$ enter the gradient flow of the diffusion objective,
and the cluster structure is learned end-to-end together with the denoiser rather than as a detached auxiliary module.

The downstream layers operate on hard cluster identities:
they consume an attention mask built from $c_\ell \in \{1, \dots, K\}$
that restricts attention to within-cluster tokens.
We score each token-subspace pair by
\begin{equation}
    r_{\ell,k} \;=\; \lVert \bm{p}_{k,\ell} \rVert
                \;=\; \lVert \bm{\mu}_k^{\top}\, \bm{x}_\ell \rVert,
    \qquad
    c_\ell \;=\; \arg\max_k r_{\ell,k},
\label{eq:hard-routing}
\end{equation}
which reuses the bilinear quantity that drives the soft aggregation.
The diagonal of $\mathbf{A}_k$ already contains $r_{\ell,k}^2 / \sqrt{h}$,
so hard routing and soft aggregation share a single geometry $\{\bm{\mu}_k\}$. 
We treat $c_\ell$ as a non-differentiable index.
Gradients to $\{\bm{\mu}_k\}$ flow exclusively through the soft path $\sum_k \mathbf{T}_k$ in Eq.~\eqref{eq:aggregation}, which in turn shapes the geometry the hard router reads off. 
The naive $\arg\max$ rule is prone to load imbalance across clusters,
and we defer mitigations to Section~\ref{sec:load_balancing}.

Overall, the module adds $K d h + 2 d^{2}$ parameters, 
the $K$ basis matrices together with the value/output projections.
Hence, introducing chunking attention into a transformer incurs no significant parameter overhead.

\subsection{Dynamic Chunking Diffusion Language Model}
\label{sec:dcdm}

The forward pass of DCDM proceeds in two stages, illustrated in Figure~\ref{fig:overview}:
(i) a chunking stage, where the input embedding is passed through a single DiT block and read by the chunking attention layer (Section~\ref{sec:cha}) to emit a per-token cluster identifier $c_\ell \in \{1, \dots, K\}$;
and (ii) a denoising stage, where the remaining DiT blocks denoise the sequence under a chunk-causal mask induced by these identifiers.

To formalize this chunking mechanism, the cluster identifiers produced in the chunking stage serve directly as the chunk indices for each position $\ell$.
The induced semantic chunks are defined as:
\begin{equation}
    \mathcal{B}_k \;=\; \bigl\{\, \ell \in \{1, \dots, L\} \;:\; c_\ell = k \,\bigr\},
    \qquad k = 1, \dots, K.
\label{eq:chunk-def}
\end{equation}
These chunks play the same role as the fixed positional blocks in BDLMs~\citep{bdlm}. However, rather than being contiguous or of uniform size, their compositions are dynamically determined per sequence by the learned centroid geometry.

Based on these semantic chunks, the denoising stage factorizes the sequence likelihood autoregressively. 
Writing $\mathbf{x}^{(k)} = (x_\ell)_{\ell \in \mathcal{B}_k}$ for the tokens of the $k$-th chunk, we have:
\begin{equation}
    p_\theta(\mathbf{x}) \;=\; \prod_{k=1}^{K}
    p_\theta\!\left( \mathbf{x}^{(k)} \mid \mathbf{x}^{(<k)} \right),
    \qquad \mathbf{x}^{(<k)} := \bigcup_{j<k} \mathbf{x}^{(j)},
\label{eq:semantic-bd}
\end{equation}
mirroring Eq.\eqref{eq:bdlm_likelihood} of BDLM but with content-defined chunks
(the derivation and proof of the corresponding DCDM NELBO objective are provided in Appendix~\ref{app:dcdm-nelbo}). 
Conceptually, each conditional is modeled as a discrete denoising diffusion process over the tokens of $\mathcal{B}_k$, governed by a base logical attention mask $\mathbf{M} \in \{0,1\}^{L \times L}$, defined as $\mathbf{M}^{\text{chunk}}_{\ell, m} = \mathbb{I}\!\left[\, c_m \le c_\ell \,\right]$.
This logical mask outlines the sequence-level topology: bidirectional attention \emph{within} a chunk and one-way attention \emph{across} chunks.

To facilitate this conditioned denoising during training, we follow the dual-stream design of BDLMs and feed DCDM a concatenated sequence $\mathbf{z}_t \oplus \mathbf{x}$.
Specifically, $\mathbf{x}^{(k)}$ is corrupted by the forward process $q(\mathbf{z}^{(k)}_t \mid \mathbf{x}^{(k)})$ at a sampled timestep $t \in (0,1]$ while $\mathbf{x}^{(<k)}$ remains clean. 
However, applying this dual-stream architecture to our chunking framework introduces a subtle risk of information leakage.
If we naively allow $\mathbf{x}$ to perform full self-attention during the chunking stage, the hidden state at each position of $\mathbf{x}$ would aggregate information from the entire clean sequence, including the very positions the denoiser must later predict.
Consequently, when $\mathbf{z}_t$ reads these $\mathbf{x}$-side hidden states for cross-chunk teacher-forcing, it would gain indirect access to its own ground-truth.

To resolve this leakage, we introduce a noise mask $\mathbf{M}^{\text{noise}}$ that mirrors the noise pattern of $\mathbf{z}_t$ onto $\mathbf{x}$, isolating the problematic positions. Let $\nu_\ell \in \{0, 1\}$ indicate whether position $\ell$ in $\mathbf{z}_t$ is currently masked; the same $\nu_\ell$ is applied to the corresponding position in the $\mathbf{x}$ half. The mask is defined as:
\begin{equation}
    \mathbf{M}^{\text{noise}}_{\ell, m}
    \;=\; \mathbb{I}[\nu_m = 0]
       \;\lor\;
       \mathbb{I}[\nu_\ell = 1 \,\land\, \ell = m].
\end{equation}
Under this noise mask, positions with a clean $\mathbf{z}_t$ counterpart ($\nu_\ell = 0$) attend to all clean keys. Conversely, positions with a masked counterpart ($\nu_\ell = 1$) can attend to all clean keys \emph{plus} themselves ($\ell = m$), but they are restricted from attending to any other masked keys. Crucially, this ensures that an $\mathbf{x}$ position corresponding to a masked token does not aggregate information from the ground-truth values of other masked positions. As a result, when the denoiser later reads $\mathbf{x}$ for teacher forcing, no backdoor path exists for the masked tokens' true values to travel back and influence the prediction.

Similarly, to accommodate this concatenated input during training, the base logical mask $\mathbf{M}^{\text{chunk}}$ must be carefully expanded. A detailed construction of the exact $2L \times 2L$ joint attention mask required for this dual-stream architecture is provided in Appendix~\ref{app:mask}.

\subsection{Load Balancing}
\label{sec:load_balancing}
 
Without explicit intervention, the hard routing of Eq.\eqref{eq:hard-routing} can collapse into severely unbalanced configurations, a failure mode extensively documented in the mixture-of-experts literature~\citep{sg_moe, scaling_moe}.
Once a centroid loses early competition, the soft path provides it with vanishingly small gradient signal, $\bm{\mu}_k$ stagnates, and the cluster never recovers. 
At a longer timescale, even when individual sequences look well-distributed, residual imbalance
accumulated over optimizer steps drives the global load away from uniform usage. 
We address these two failure modes with complementary mechanisms at two timescales: a \emph{per-sequence} auxiliary loss against intra-sequence starvation, 
and, following~\citet{aux_loss_free}, a \emph{global-batch} bias correction that stabilizes load over the batch and across training steps.
 
\paragraph{Per-sequence load balancing.}
Within a single sequence, we encourage every centroid to receive a non-trivial share of the tokens through an auxiliary loss applied to a differentiable hard sample of the routing scores.
For each token, we draw
\begin{equation}
    \tilde{\mathbf{c}}_\ell \;=\; \mathrm{GumbelSoftmax}_{\mathrm{ST}}(\bm{r}_\ell)
    \in \{0,1\}^K,
\end{equation}
a one-hot vector under the straight-through estimator~\citep{gumbel_softmax},
whose forward value is
a hard sample but whose backward path follows the softmax surrogate and thus
carries gradients into $\{\bm{\mu}_k\}$. With per-sequence usage frequency
$f_{b,k} = \tfrac{1}{L}\sum_{\ell=1}^{L} [\tilde{\mathbf{c}}_\ell]_k$ for
sequence $b$ in a batch of size $B$, we minimize
\begin{equation}
    \mathcal{L}_{\mathrm{chunk}}
    \;=\; - \frac{1}{B K} \sum_{b=1}^{B} \sum_{k=1}^{K}
            \log\bigl(f_{b,k} + \varepsilon\bigr),
\label{eq:chunk-loss}
\end{equation}
where $\varepsilon$ is a small positive constant.
Under the simplex constraint $\sum_k f_{b,k} = 1$, the loss attains its minimum when every sequence distributes its tokens uniformly across the $K$ clusters.
Because $\mathcal{L}_{\mathrm{chunk}}$ is computed independently per sequence,
it acts on the fastest available timescale, a single forward pass, and prevents any cluster from being starved within an individual example.
 
\paragraph{Global-batch load balancing.}
Even when each sequence uses all $K$ clusters, the overall load can still drift over the batch and across training steps. 
Therefore, we add a non-trainable per-cluster bias $\mathbf{b} \in \mathbb{R}^K$ to the
routing scores at the hard-assignment step only:
\begin{equation}
    c_\ell \;=\; \arg\max_k\, \bigl(r_{\ell,k} + b_k\bigr),
\end{equation}
while the soft aggregation in Eq.\eqref{eq:aggregation} continues to use the unbiased scores. 
We track a running token count
$N_k = \sum_\ell \mathbb{I}[c_\ell = k]$ for each centroid (with total $N = \sum_k N_k$), and update the bias once per update interval by
\begin{equation}
    b_k \;\leftarrow\; b_k - \eta_b \,\bigl( N_k / N - 1/K \bigr),
\end{equation}
where $\eta_b > 0$ is a step size controlling how aggressively the bias reacts to load imbalance.
We use a fixed $\eta_b$ throughout training. 
The update lowers the bias of overloaded clusters and raises that of underused ones.
Because $\mathbf{b}$ enters only the discrete $\arg\max$ branch, no gradient flows through it: 
the soft path that trains $\{\bm{\mu}_k\}$ is unaffected, and $\mathbf{b}$ acts purely as a control-loop correction layered on top of the resulting hard assignments.


\section{Experiments}
\label{sec:expt}
 
\subsection{Experimental Setup}

\begin{wrapfigure}{r}{0.45\textwidth}
    \vspace{0pt}
    \centering
        \includegraphics[width=\linewidth]{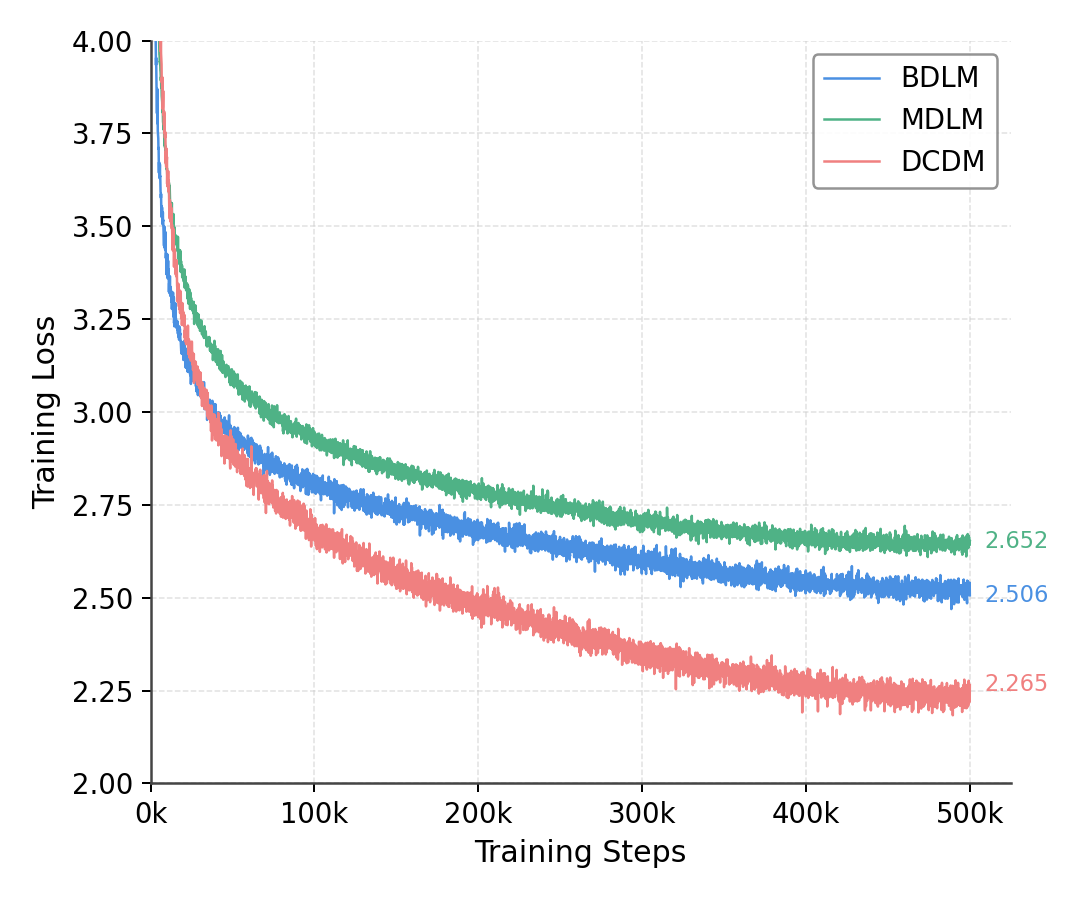}
        \caption{
            Training loss against training steps for the three dense
            diffusion models. Final-step values confirm the ordering
            observed on downstream metrics.
        }
    \label{fig:training_loss}
    \vspace{-20pt}
\end{wrapfigure}
 
\paragraph{Datasets.}
All diffusion models are pretrained on OpenWebText~\cite{openwebtext} under a unified training protocol; full training details are deferred to Appendix~\ref{app:appendix-impl}. 
We then evaluate the pretrained LLMs on a suite of standard benchmarks
grouped into three categories: general reasoning and knowledge (ARC-C~\cite{arc},MMLU~\cite{mmlu_1, mmlu_2}, HellaSwag~\cite{hellaswag},  TruthfulQA~\cite{truthfulqa}, WinoGrande~\cite{winogrande}, PIQA~\cite{piqa}), mathematical reasoning (MATH~\cite{math}, GSM8K~\cite{gsm8k}), and code generation (HumanEval~\cite{humaneval}). 
Training dynamics for the dense diffusion models are summarized in
Figure~\ref{fig:training_loss}. 
A complementary zero-shot language-modeling evaluation on seven held-out corpora is reported in Appendix~\ref{app:zero-shot-ppl}.
 
\paragraph{Baselines.}
We compare DCDM against the dominant paradigms for discrete diffusion
language modeling.
(i) \textbf{MDLM}~\citep{mdlm} is a masked discrete diffusion language
model that denoises tokens in parallel without block structure.
(ii) \textbf{BDLM}~\citep{bdlm} is a block discrete diffusion model that
imposes an autoregressive factorization over fixed-size \emph{positional}
blocks, and is the closest prior work to ours.
(iii) \textbf{AdaBlock-dLLM}~\citep{adablock} is a training-free inference-time
scheduler that adapt block size during decoding on top of a model
trained with fixed positional blocks.
We evaluate all baselines at matched parameter scale, and additionally
include a sparse \textbf{DCDM\,(MoE)} variant whose active-parameter
budget (0.4B / 1.2B) is slightly below the dense models, to assess
whether the semantic chunk structure of DCDM composes with conditional
computation. All dense diffusion models are evaluated at two scales,
0.5B and 1.5B parameters, with identical training data and tokenizer,
under our unified evaluation protocol.
 
\paragraph{Metrics.}
For the downstream benchmarks we follow each task's standard metric~\cite{diffugpt}:
accuracy for the multiple-choice tasks (ARC-C, MMLU, HellaSwag, TruthfulQA, WinoGrande, PIQA), exact-match accuracy for MATH and GSM8K, and pass@1 for HumanEval. The number of in-context examples used for each benchmark is reported in parentheses in Table~\ref{tab:benchmark_pretrain}; 
entries without a parenthesis are evaluated zero-shot.
 
\begin{table}[t]
    \centering
    \caption{Downstream benchmark results for pretrained LLMs at the 0.5B and 1.5B scales. Numbers in parentheses indicate the number of in-context examples; entries without parentheses are zero-shot. $\dagger$ marks benchmarks for which the model has been fine-tuned on the corresponding training split. For DCDM\,(MoE), the \textbf{Active Params} row reports activated parameters per token with total parameters in parentheses; all other models activate all parameters at every token. Within each scale, the best result per row is shown in \textbf{bold} and the second-best is \underline{underlined}. $\ast$ Average is computed over all nine reported benchmarks.}
    \label{tab:benchmark_pretrain}
    \resizebox{\textwidth}{!}{
    \begin{tabular}{l|cccc|cccc}
        \toprule
        \multirow{2}{*}{\textbf{Model}}
        & \multicolumn{4}{c}{\textbf{0.5B Scale}}
        & \multicolumn{4}{c}{\textbf{1.5B Scale}} \\
        \cmidrule(lr){2-5} \cmidrule(lr){6-9}
        & MDLM & BDLM & DCDM & DCDM\,(MoE)
        & MDLM & BDLM & DCDM & DCDM\,(MoE) \\
        \textbf{Active Params}
        & 0.5B & 0.5B & 0.5B & 0.4B (0.8B)
        & 1.5B & 1.5B & 1.5B & 1.2B (2.8B) \\
        \midrule
        ARC-C
        & 23.29 & 23.21 & \underline{24.49} & \textbf{24.74}
        & 26.79 & 26.71 & \underline{27.13} & \textbf{27.50} \\
        MMLU
        & 23.24 & 23.53 & \underline{24.32} & \textbf{25.09}
        & 24.49 & 25.22 & \underline{25.59} & \textbf{26.10} \\
        HellaSwag
        & 31.31 & 37.52 & \textbf{40.90} & \underline{40.61}
        & 42.81 & 50.66 & \underline{51.04} & \textbf{51.28} \\
        TruthfulQA
        & 41.43 & 39.85 & \underline{42.65} & \textbf{43.86}
        & 42.17 & 39.59 & \underline{44.16} & \textbf{44.80} \\
        WinoGrande
        & 49.04 & \textbf{52.35} & 49.88 & \underline{51.30}
        & \underline{53.20} & \textbf{54.06} & 52.80 & 52.85 \\
        PIQA
        & 57.13 & 58.81 & \underline{58.87} & \textbf{59.68}
        & 61.21 & 65.23 & \underline{66.05} & \textbf{66.32} \\
        \midrule
        MATH (2-shot)$^\dagger$
        & 7.40  & 9.60  & \underline{10.20} & \textbf{11.20}
        & 18.80 & 21.40 & \underline{22.60} & \textbf{23.20} \\
        GSM8K (4-shot)$^\dagger$
        & 45.32 & 50.21 & \underline{50.37} & \textbf{50.62}
        & 53.55 & 57.13 & \underline{58.72} & \textbf{59.40} \\
        \midrule
        HumanEval
        & 0.48  & 1.16  & \underline{1.45}  & \textbf{1.55}
        & 2.23  & 2.81  & \textbf{3.01}  & \underline{2.97} \\
        \midrule
        Average$^{\ast}$
        & 30.96 & 32.92 & \underline{33.68} & \textbf{34.29}
        & 36.14 & 38.09 & \underline{39.01} & \textbf{39.38} \\
        \bottomrule
    \end{tabular}%
    }
\end{table}
 
\subsection{Main Results}
 
\paragraph{Downstream benchmarks.}
Table~\ref{tab:benchmark_pretrain} reports
zero- and few-shot results across
the downstream suite at both 0.5B and 1.5B scales. Three patterns hold
consistently across scales.
 
First, DCDM outperforms the unstructured diffusion baseline MDLM on every
benchmark except WinoGrande, where the two are within 1 point at both
scales (DCDM ahead by 0.84 at 0.5B, behind by 0.40 at 1.5B). The largest
gains are concentrated on HellaSwag ($+9.59$ / $+8.23$ points at 0.5B / 1.5B),
GSM8K ($+5.05$ / $+5.17$), and PIQA ($+1.74$ / $+4.84$); smaller but
consistent gains of $0.3$--$2.0$ points appear on ARC-C, MMLU, TruthfulQA,
and HumanEval. This indicates that the semantic chunk factorization
introduced in Section~\ref{sec:dcdm} provides a clear inductive advantage
over parallel denoising without block structure.
 
Second, DCDM also surpasses the positional-block baseline BDLM on every
benchmark except WinoGrande, where it lags by 2.47 points at 0.5B and
1.26 points at 1.5B. Since BDLM shares the same block-autoregressive
factorization but uses fixed positional blocks, this gap is attributable to
the content-defined chunk partition of DCDM rather than to the block
structure itself.
 
Third, DCDM\,(MoE) yields a small but consistent improvement over dense
DCDM, averaging $+0.57$ points at 0.5B and $+0.33$ points at 1.5B across the
suite, at a slightly reduced active-parameter budget (0.4B vs.\ 0.5B; 1.2B
vs.\ 1.5B). The gain is modest but indicates that the semantic chunk
structure of DCDM composes cleanly with sparse conditional computation
rather than interfering with it.

Relative orderings among the four models are largely preserved when scaling
from 0.5B to 1.5B: DCDM\,(MoE) typically leads the suite, followed by dense
DCDM, with WinoGrande the consistent exception in favor of BDLM. The size
of DCDM's advantage over the diffusion baselines is benchmark-dependent.
At the suite level, the overall gap between DCDM
and the diffusion baselines is broadly preserved across scales.

\begin{table}[t]
    \centering
    \caption{
        Comparison against alternative block-partitioning strategies at the 1.5B scale on math (MATH, GSM8K) and code (HumanEval) benchmarks. Models are grouped by the type of block structure used at sampling time. 
        Best per row in \textbf{bold}, second-best \underline{underlined}. $\ast$\,Average is computed over the three reported benchmarks.
    }
    \label{tab:benchmark_adaptive}
    \begin{tabular}{l|ccccc}
        \toprule
        \textbf{Model}
        & \textbf{None} & \textbf{Fix}
        & \textbf{Adaptive}
        & \multicolumn{2}{c}{\textbf{Dynamic}} \\
        \cmidrule(lr){2-2} \cmidrule(lr){3-3} \cmidrule(lr){4-4} \cmidrule(lr){5-6}
        & MDLM & BDLM & AdaBlock & DCDM & DCDM\,(MoE) \\
        \textbf{Active Params}
        & 1.5B & 1.5B & 1.5B & 1.5B & 1.2B (2.8B) \\
        \midrule
        MATH (2-shot)$^\dagger$
        & 18.80 & 21.40 & 21.20 & \underline{22.80} & \textbf{23.20} \\
        GSM8K (4-shot)$^\dagger$
        & 53.55 & 57.16 & 57.48 & \underline{58.72} & \textbf{59.40} \\
        \midrule
        HumanEval (0-shot)
        & 2.23  & 2.81  & \textbf{3.01}  & \textbf{3.01}  & 2.97 \\
        \midrule
        Average$^{\ast}$
        & 24.86 & 27.12 & 27.23 & \underline{28.17} & \textbf{28.52} \\
        \bottomrule
    \end{tabular}%
\end{table}
 
\paragraph{Comparison against adaptive block-partitioning methods.}
We further compare DCDM against AdaBlock, two recent methods that adapt block size \emph{at sampling time} on top of a model trained with fixed positional blocks.
As shown in Table~\ref{tab:benchmark_adaptive},
DCDM attains the best dense average ($28.17$), clearly above AdaBlock ($27.23$); DCDM\,(MoE) further improves to $28.52$ at a smaller active-parameter budget ($1.2$B vs.\ $1.5$B).
Beyond the headline numbers, these adaptive baselines adjust block size only at sampling time, so the partition seen at inference is not the one the model was optimized against; DCDM removes this train-test mismatch by design.

\begin{figure*}[t]
    \centering
    \begin{minipage}[t]{0.48\textwidth}
        \centering
        \includegraphics[width=\linewidth]{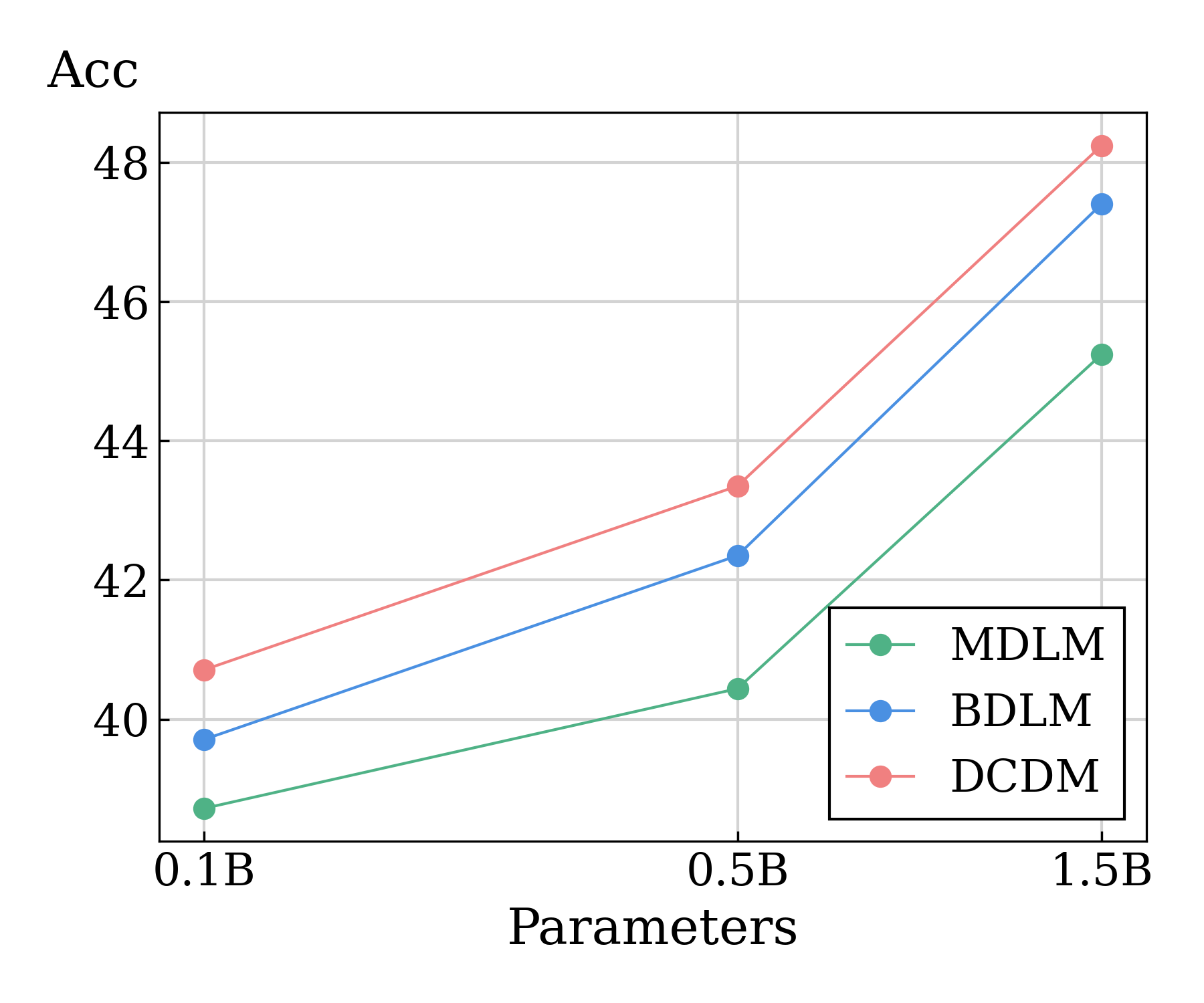}
        \subcaption{Parameters vs.\ accuracy.}
        \label{fig:params_vs_acc}
    \end{minipage}\hfill
    \begin{minipage}[t]{0.48\textwidth}
        \centering
        \includegraphics[width=\linewidth]{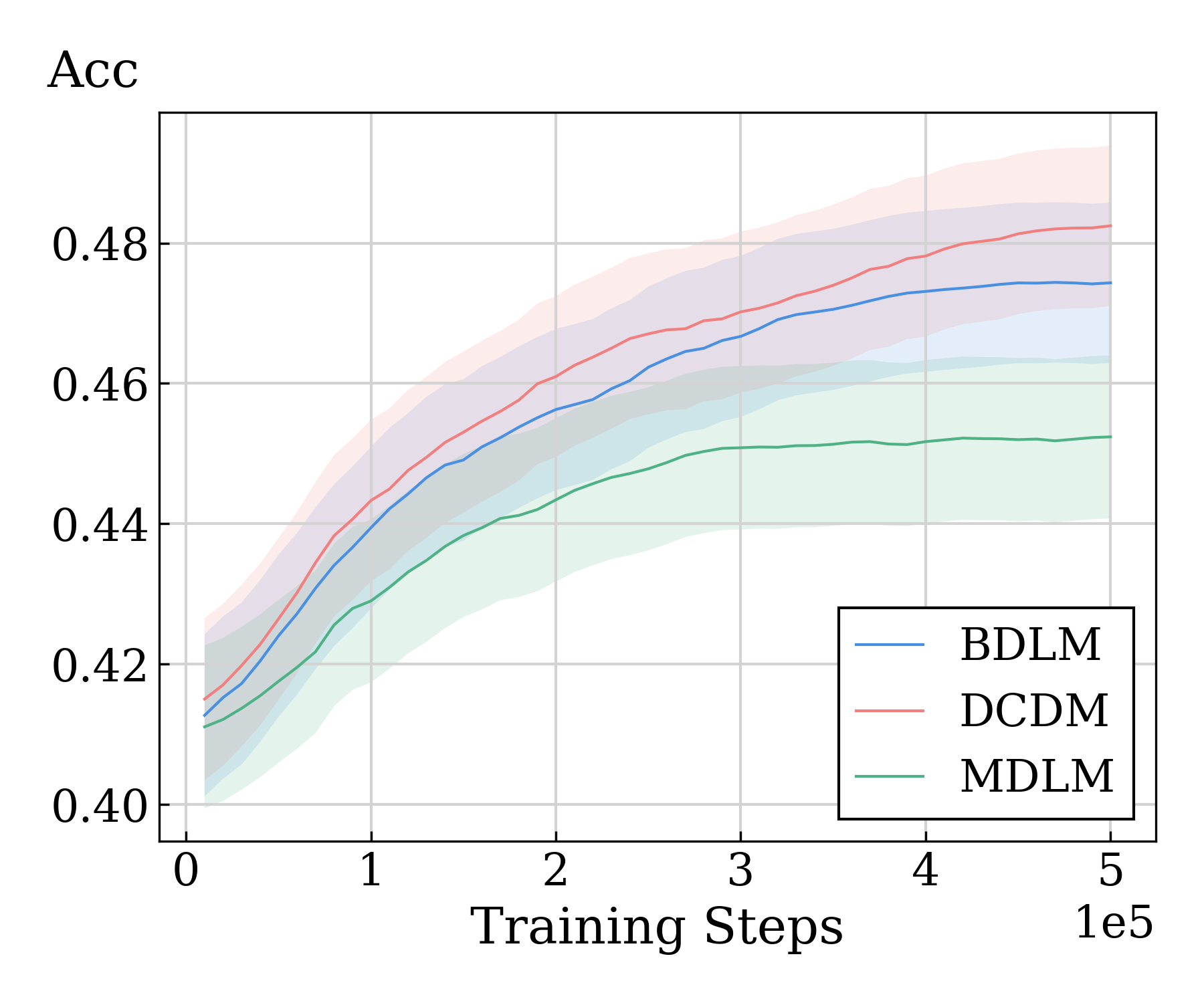}
        \subcaption{Training steps vs.\ accuracy.}
        \label{fig:steps_vs_acc}
    \end{minipage}
    \caption{
        Scaling and training efficiency of the three dense diffusion models (MDLM, BDLM, DCDM) on the downstream benchmark suite.
        \textbf{(a)} Suite-average accuracy at three parameter scales (0.1B, 0.5B, 1.5B), each trained under our standard token budget.
        \textbf{(b)} Suite-average accuracy along the training trajectory at the 1.5B scale; shaded regions denote one standard deviation across runs.
    }
    \label{fig:scaling_efficiency}
\end{figure*}
 
\paragraph{Scaling and Training.}
Figure~\ref{fig:scaling_efficiency} examines how the three dense diffusion
models trade off accuracy against (a) parameter count and (b) the number of
training steps consumed.
 
Figure~\ref{fig:params_vs_acc} reports suite-average accuracy at three
parameter scales. The ordering $\text{DCDM} > \text{BDLM} > \text{MDLM}$
holds at every scale we evaluate. The gap between MDLM and the two
block-structured models widens with scale, indicating that block structure
is not merely a small-scale regularizer but contributes more as capacity
grows. The gap between DCDM and BDLM is smaller throughout but remains
positive across scales, isolating the contribution of content-defined, as
opposed to merely positional, blocks once the model is large enough to
exploit either form of partition.
 
Figure~\ref{fig:steps_vs_acc} traces accuracy along the training trajectory
at the 1.5B scale. 
The three models start from comparable accuracy, but the ordering $\text{DCDM} > \text{BDLM} > \text{MDLM}$ emerges within the early portion of training and remains stable thereafter.
MDLM plateaus earliest and at the lowest level, while BDLM and DCDM continue to improve toward the end of the run. 
DCDM matches MDLM's final accuracy well before MDLM finishes training, and reaches BDLM's final accuracy noticeably earlier than BDLM does, indicating genuinely faster optimization rather than convergence to a similar asymptote along a different path. 
The reason is structural:
routing tokens into content-defined blocks supplies denoising targets that are coherent at the level of the partition, so the gradient signal at each step concentrates on within-block dependencies rather than being diluted across unrelated positions. 

\begin{table}[t]
\centering
\caption{Per-benchmark results for the centroid-count ablation
($K \in \{8, 16, 32, 64\}$) at the 0.5B scale. Best in each column in
\textbf{bold}.}
\label{tab:k-ablation-full}
\begin{tabular}{l|cccccc|c}
\toprule
$K$    & ARC-C & MMLU  & HellaSwag & TruthfulQA & WinoGrande & PIQA  & Average \\
\midrule
$K=8$  & \underline{24.49} & 23.32 & \textbf{40.90} & 42.65 & \textbf{49.88} & 58.87 & 40.01 \\
$K=16$ & 23.55 & \textbf{24.75} & \underline{40.47} & \textbf{43.91} & 49.25 & \textbf{60.61} & \textbf{40.58} \\
$K=32$ & 24.32 & \underline{23.89} & 39.31 & 43.35 & \underline{49.75} & \underline{60.01} & \underline{40.11} \\
$K=64$ & \textbf{24.83} & 23.40 & 39.08 & \underline{43.78} & 49.28 & 58.98 & 40.06 \\
\bottomrule
\end{tabular}
\end{table}

\subsection{Ablation Study: Number of Clusters}
A central hyperparameter of DCDM is the number of clusers $K$ in the chunking attention layer, which determines the maximum granularity of the semantic partition. 
We isolate its effect with an ablation at the 0.5B scale, training DCDM with $K \in \{8, 16, 32, 64\}$ under otherwise identical settings and evaluating on the general-reasoning subset of the downstream benchmark suite. 
Per-benchmark numbers are reported in Table~\ref{tab:k-ablation-full}.

On the suite average, $K=16$ achieves the best result ($40.58$),
ahead of the alternatives by a modest but consistent margin ($K=8$: $40.01$; $K=32$: $40.11$; $K=64$: $40.06$).
The trend is non-monotonic and exhibits a clear interior optimum:
average performance peaks at $K=16$ and degrades at both extremes.
This U-shape is consistent with the two failure modes that bracket the choice of $K$:
too few clusers under-partition the sequence and weaken the selectivity of the bilinear gate of Eq.~\eqref{eq:bilinear-gate}, pushing the soft aggregation toward the unstructured limit of MDLM;
too many clusers over-fragment the partition,
leaving each cluster with too few tokens to support meaningful intra-block bidirectional denoising and forcing the load-balancing mechanisms of Section~\ref{sec:load_balancing} to operate near their stability margin.


\section{Conclusion}

We introduced DCDM, a discrete diffusion language model that replaces the fixed positional blocks of BDLM with content-defined chunks produced by a learned subspace-clustering attention layer. 
DCDM matches BDLM in generation flexibility while consistently improving training loss, downstream accuracy, and zero-shot perplexity at every scale we tested, and it admits MoE
extensions for further gains.


\bibliographystyle{plainnat}
\bibliography{refs}


\newpage
\appendix

\section{Pseudocode}
\label{app:pseudocode}

Algorithm~\ref{alg:chunking-attention} reproduces the chunking attention layer of Section~\ref{sec:cha} as a single-batch computation.
The notation follows the main text: $L$ is the sequence length, $d$ the model dimension, $K$ the number of clusters, and $h$ the per-cluster subspace dimension.
The load-balancing bias $\mathbf{b} \in \mathbb{R}^{K}$ is the running correction defined in Section~\ref{sec:load_balancing}; it is updated externally once per
optimizer interval and carries no gradient.

\begin{algorithm}
\caption{Chunking Attention (single batch element).}
\label{alg:chunking-attention}
\begin{algorithmic}[1]
\Require input embedding $\mathbf{H} \in \mathbb{R}^{L \times d}$;
         centroid projections
         $\{\bm{\mu}_k\}_{k=1}^{K}$, $\bm{\mu}_k \in \mathbb{R}^{d \times h}$;
         value/output projections
         $\mathbf{W}_V, \mathbf{W}_O \in \mathbb{R}^{d \times d}$;
         load-balancing bias $\mathbf{b} \in \mathbb{R}^{K}$
\Ensure   updated embedding $\mathbf{Y} \in \mathbb{R}^{L \times d}$;
          hard cluster ids $\mathbf{c} \in \{1,\dots,K\}^{L}$;
          per-sequence chunk loss $\mathcal{L}_{\mathrm{chunk}}$
\vspace{2pt}
\Statex \textit{// Soft path: project, score, softmax, aggregate.}
\State $\bm{p}_{k,\ell} \gets \bm{\mu}_k^{\top}\, \bm{x}_\ell$
       \Comment{$\bm{p}_{k,\ell} \in \mathbb{R}^{h}$ for all $\ell, k$}
\For{$k = 1, \dots, K$}
    \State $[\mathbf{A}_k]_{\ell, m} \gets
            \bm{p}_{k,\ell}^{\!\top}\, \bm{p}_{k,m} / \sqrt{h}$
    \State $\mathbf{S}_k \gets \mathrm{softmax}(\mathbf{A}_k)$
           \Comment{row-wise; $\mathbf{S}_k \in \mathbb{R}^{L \times L}$}
\EndFor
\State $\mathbf{Y} \gets \mathbf{W}_O \!\Big(
        \tfrac{1}{\sqrt{K}} \textstyle\sum_{k} \mathbf{S}_k
       \Big)\! \mathbf{W}_V\, \mathbf{H}$
\vspace{2pt}
\Statex \textit{// Hard routing: read off the cluster id used by the mask.}
\State $r_{\ell,k} \gets \lVert \bm{p}_{k,\ell} \rVert$
       \Comment{reuses the diagonal of $\mathbf{A}_k$}
\State $c_\ell \gets \arg\max_{k}\, ( r_{\ell,k} + b_k )$
       \Comment{$\mathbf{b}$ updated externally, no gradient}
\vspace{2pt}
\Statex \textit{// Per-sequence load-balancing loss.}
\State $\tilde{\mathbf{c}}_\ell \gets
       \mathrm{GumbelSoftmax}_{\mathrm{ST}}(\bm{r}_\ell)$
       \Comment{$\bm{r}_\ell = (r_{\ell,1},\dots,r_{\ell,K})$;
                differentiable hard sample}
\State $f_k \gets \tfrac{1}{L} \textstyle\sum_{\ell} [\tilde{\mathbf{c}}_\ell]_k$
       \Comment{per-sequence usage frequency of centroid $k$}
\State $\mathcal{L}_{\mathrm{chunk}} \gets
       -\tfrac{1}{K} \textstyle\sum_{k} \log(f_k + \varepsilon)$
\vspace{2pt}
\State \Return $\mathbf{Y},\, \mathbf{c},\, \mathcal{L}_{\mathrm{chunk}}$
\end{algorithmic}
\end{algorithm}

The algorithm splits into three logical blocks. 
The \emph{soft path} projects $\mathbf{H}$ onto every cluster subspace, 
forms the per-centroid affinity matrices $\mathbf{A}_k$ and their row-wise softmaxes $\mathbf{T}_k$, and aggregates them into the module output $\mathbf{Y}$;
this is the only block that carries gradients to the clusers $\{\bm{\mu}_k\}$.
The \emph{hard routing} block reads off the cluster identity $c_\ell$ from the alignment scores $r_{\ell,k}$ after adding the load-balancing bias $\mathbf{b}$; 
the resulting $\mathbf{c}$ is the index the downstream attention mask $\mathbf{M}^{\text{chunk}}$ is built from.
The \emph{load-balancing} block draws a straight-through Gumbel-softmax sample $\tilde{\mathbf{c}}_\ell$ and uses it to compute the per-sequence usage frequency $f_k$ and the auxiliary loss $\mathcal{L}_{\mathrm{chunk}}$, 
which back-propagates into $\{\bm{\mu}_k\}$ alongside the diffusion objective.

\section{Attention Mask Construction}
\label{app:mask}
 
\paragraph{Noise mask.}
Let $\nu_\ell \in \{0, 1\}$ indicate whether position $\ell$ (taken modulo $L$ on the doubled sequence) is a masked token at the current diffusion timestep. 
The noise mask is
\begin{equation}
    \mathbf{M}^{\text{noise}}_{\ell, m}
    \;=\; \mathbb{I}[\nu_m = 0]
       \;\lor\;
       \mathbb{I}[\nu_\ell = 1 \,\land\, \ell = m].
\end{equation}
A clean query attends only to clean keys; a noisy query attends to all clean keys plus its own position. 
The first clause prevents the information-less mask-token embedding from contaminating the representations of clean tokens, 
and the second clause is a self-loop that keeps each noisy position identifiable through the layer.
 
\paragraph{Chunk mask.}
At inference time the chunk mask reduces directly to the chunk-causal
mask,
\begin{equation}
    \mathbf{M}^{\text{chunk}}_{\ell, m}
    \;=\; \mathbb{I}[c_m \le c_\ell],
\end{equation}
where $c_\ell$ is the cluster index of position $\ell$. At training
time, the doubled sequence requires three explicit cases. Let
$\sigma_\ell \in \{0, 1\}$ indicate whether position $\ell$ lies in the
noisy half ($\sigma_\ell = 1$ iff $\ell < L$), and let $c_\ell$ denote
the cluster index of position $\ell \bmod L$. The training-time chunk
mask is
\begin{equation}
\begin{aligned}
    \mathbf{M}^{\text{chunk}}_{\ell, m}
    \;=\;\;
    & \mathbb{I}[\sigma_\ell = 1,\, \sigma_m = 1,\, c_\ell = c_m] \\
    \lor\;\; & \mathbb{I}[\sigma_\ell = 1,\, \sigma_m = 0,\, c_\ell > c_m] \\
    \lor\;\; & \mathbb{I}[\sigma_\ell = 0,\, \sigma_m = 0,\, c_\ell \ge c_m].
\end{aligned}
\end{equation}
The three clauses cover, respectively: bidirectional attention within a chunk in the noisy half (parallel denoising of co-clustered tokens);
attention from a noisy query to a clean key in a strictly earlier chunk (cross-block conditioning, sourced from the clean half);
and the standard chunk-causal mask within the clean half (autoregressive
teacher-forcing).
Notably, no clause connects a clean query to a noisy key, so the autoregressive teacher-forcing signal is never contaminated by noise.

\section{DCDM NELBO}
\label{app:dcdm-nelbo}
 
We provide the NELBO derivation for DCDM.
The derivation follows BDLMs~\citep{bdlm} step for step; 
the only substantive change is that DCDM factorizes the joint distribution over the $K$ \emph{semantic} chunks $\{\mathcal{B}_k\}_{k=1}^{K}$ produced by the chunking attention module of Section~\ref{sec:cha}, 
with the cluster identifier $c_\ell$ used directly as the chunk index, rather than over fixed-size positional blocks.
 
Let $\mathbf{x}_{1:L} = [\mathbf{x}_1, \ldots, \mathbf{x}_L]$ be a sequence drawn from the data distribution $q(\mathbf{x})$. Following the partition of Section~\ref{sec:dcdm}, we set
\begin{equation*}
\mathbf{x}^{(k)} \;:=\; (\mathbf{x}_\ell)_{\ell \in \mathcal{B}_k},
\qquad
\mathbf{x}^{(<k)} \;:=\; \bigcup_{j<k} \mathbf{x}^{(j)},
\qquad k \in \{1, \ldots, K\},
\end{equation*}
with block size $L_k := |\mathcal{B}_k|$. Unlike the positional blocks of
BDLM, $L_k$ is not uniform; it is determined per sequence by the learned
routing. For brevity we abbreviate $\mathbf{x}^{(k)}$ to $\mathbf{x}^k$
throughout this appendix. Each block undergoes diffusion over $T$
discretization steps, with $t(i) = i/T$ and $s(i) = (i-1)/T$ for
$i \in [1, T]$; let $\mathrm{D}_{\mathrm{KL}}[\,\cdot\,]$ denote the
Kullback--Leibler divergence. Then
\begin{align}
-\log p_\theta(\mathbf{x})
&= -\sum_{k=1}^{K} \log p_\theta\bigl(\mathbf{x}^k \,\big|\, \mathbf{x}^{<k}\bigr) \notag \\
&= -\sum_{k=1}^{K} \log \mathbb{E}_q
   \frac{p_\theta\bigl(\mathbf{x}^k_{t(1):t(T)} \,\big|\, \mathbf{x}^{<k}\bigr)}
        {q\bigl(\mathbf{x}^k_{t(1):t(T)} \,\big|\, \mathbf{x}^k\bigr)} \notag \\
&= -\sum_{k=1}^{K} \log \mathbb{E}_q
   \frac{p_\theta\bigl(\mathbf{x}^k_{t(T)} \,\big|\, \mathbf{x}^{<k}\bigr)
         \prod_{i=1}^{T} p_\theta\bigl(\mathbf{x}^k_{s(i)} \,\big|\, \mathbf{x}^k_{t(i)}, \mathbf{x}^{<k}\bigr)}
        {\prod_{i=1}^{T} q\bigl(\mathbf{x}^k_{t(i)} \,\big|\, \mathbf{x}^k_{s(i)}\bigr)} \notag \\
&\leq \sum_{k=1}^{K} \Big[
   \underbrace{-\mathbb{E}_q \log p_\theta\bigl(\mathbf{x}^k \,\big|\, \mathbf{x}^k_{t = 1/T}, \mathbf{x}^{<k}\bigr)}_{\mathcal{L}_{\text{recons}}}
   \notag \\
&\quad + \underbrace{\mathbb{E}_{t \in \{2/T, \ldots, (T-1)/T, 1\}} \mathbb{E}_q\, T\,
   \mathrm{D}_{\mathrm{KL}}\!\bigl(q(\mathbf{x}^k_s | \mathbf{x}^k_t, \mathbf{x}^k)
                                \,\big\|\,
                                p_\theta(\mathbf{x}^k_s | \mathbf{x}^k_t, \mathbf{x}^{<k})\bigr)}_{\mathcal{L}_{\text{diffusion}}}
   \notag \\
&\quad + \underbrace{\mathrm{D}_{\mathrm{KL}}\!\bigl(q(\mathbf{x}^k_{t=1} | \mathbf{x}^k)
                                                  \,\big\|\,
                                                  p_\theta(\mathbf{x}^k_{t=1})\bigr)}_{\mathcal{L}_{\text{prior}}}
\Big]. \label{eq:dcdm-nelbo}
\end{align}
 
We now specialize Eq.\eqref{eq:dcdm-nelbo} to the masked diffusion process used in our implementation.
In BDLM, the denoiser conditions on positionally earlier blocks;
in DCDM, it conditions on semantically earlier chunks $\mathbf{x}^{<k}$, with the conditioning realized operationally by the chunk-causal attention mask $\mathbf{M}^{\text{chunk}}$. 
Since the SUBS parameterization of~\citet{mdlm} constrains $p_\theta$ token-wise, enforcing zero masking probabilities and carry-over unmasking on a per-token basis,
it does not depend on the block structure, and its derivation carries over verbatim once the variable-length DCDM chunks $\mathcal{B}_k$ of size $L_k$ are substituted for the fixed-size positional blocks of BDLM.
 
Following~\citet{mdlm}, the diffusion loss simplifies to
\begin{align}
\mathcal{L}_{\text{diffusion}}
&= \sum_{k=1}^{K} \mathbb{E}_t \mathbb{E}_q\, T \left[
    \sum_{\ell=1}^{L_k}
    \mathrm{D}_{\mathrm{KL}}\!\bigl(
        q(\mathbf{x}^{k,\ell}_s | \mathbf{x}^{k,\ell}_t, \mathbf{x}^{k,\ell})
        \,\big\|\,
        p_\theta(\mathbf{x}^{k,\ell}_s | \mathbf{x}^k_t, \mathbf{x}^{<k})
    \bigr)
\right] \notag \\
&= \sum_{k=1}^{K} \mathbb{E}_t \mathbb{E}_q\, T \left[
    \frac{\alpha_t - \alpha_s}{1 - \alpha_t}\,
    \log p_\theta\bigl(\mathbf{x}^k \,\big|\, \mathbf{x}^k_t, \mathbf{x}^{<k}\bigr)
\right].
\end{align}
 
Taking $T \to \infty$ with $T(\alpha_t - \alpha_s) = \alpha'_t$ yields the
continuous-time form
\begin{equation}
\mathcal{L}_{\text{diffusion}}
\;=\; \sum_{k=1}^{K} \mathbb{E}_{t \sim [0,1]} \mathbb{E}_q
   \left[
       \frac{\alpha'_t}{1 - \alpha_t}\,
       \log p_\theta\bigl(\mathbf{x}^k \,\big|\, \mathbf{x}^k_t, \mathbf{x}^{<k}\bigr)
   \right].
   \label{eq:dcdm-diffusion-cts}
\end{equation}
By the same arguments as in~\citet[Suppl.~A.2.4]{mdlm},
$\mathcal{L}_{\text{recons}} = 0$ in the continuous-time limit because
$\mathbf{x}^k_{t(1)} \sim \lim_{T \to \infty} \mathrm{Cat}(\,\cdot\,;\,\mathbf{x}^k_{t = 1/T})
 = \mathrm{Cat}(\,\cdot\,;\,\mathbf{x}^k)$,
and $\mathcal{L}_{\text{prior}} = 0$ because $\alpha_{t=1} = 0$ ensures
$q(\mathbf{x}^k_{t=1} | \mathbf{x}^k) = \mathrm{Cat}(\,\cdot\,;\,\mathbf{m})
 = p_\theta(\mathbf{x}^k_{t=1})$. Consequently, the final DCDM diffusion
objective is
\begin{equation}
\mathcal{L}_{\text{DCDM}}(\mathbf{x}; \theta)
\;=\; \sum_{k=1}^{K} \mathbb{E}_{t \sim [0,1]} \mathbb{E}_q
   \left[
       \frac{\alpha'_t}{1 - \alpha_t}\,
       \log p_\theta\bigl(\mathbf{x}^k \,\big|\, \mathbf{x}^k_t, \mathbf{x}^{<k}\bigr)
   \right],
   \label{eq:dcdm-loss}
\end{equation}
which is invariant to the choice of noise schedule
$\alpha_t$~\citep[Suppl.~E.1.1]{mdlm}.

\section{Implementation Details}
\label{app:appendix-impl}

\subsection{Training Setup}

All models are trained from scratch with the AdamW optimizer ($\beta_1{=}0.9$, $\beta_2{=}0.999$, $\epsilon{=}10^{-8}$, weight decay $0.01$) using a cosine learning-rate schedule with $2{,}500$ linear-warmup steps and a peak learning rate of $3\times 10^{-4}$.
The global training batch size and compute footprint scale with model size: $0.1$B models use a global batch size of $128$ on $8 \times$ H800 GPUs, while models of $0.5$B and above use a global batch size of $512$ on $64 \times$ H800 GPUs.
Gradients are clipped to a maximum norm of $1.0$. We train for $500{,}000$ optimization steps in \texttt{bfloat16} mixed precision with TF32 matmul allowed.
Diffusion training uses a time-sampling lower bound of $\epsilon_t{=}10^{-3}$.
For DCDM variants the chunking auxiliary loss is weighted by $10^{-2}$, and for the MoE variants the router auxiliary loss is also weighted by $10^{-2}$. 
A summary of the optimizer/training hyperparameters is given in
Table~\ref{tab:train-hparams}.

\begin{table}[h]
\centering
\small
\caption{Training hyperparameters shared across all models.}
\label{tab:train-hparams}
\begin{tabular}{ll}
\toprule
\textbf{Hyperparameter} & \textbf{Value} \\
\midrule
Optimizer                  & AdamW \\
$(\beta_1, \beta_2, \epsilon)$ & $(0.9,\ 0.999,\ 10^{-8})$ \\
Weight decay               & $0.01$ \\
Peak learning rate         & $3\times 10^{-4}$ \\
LR schedule                & cosine \\
Warmup steps               & $2{,}500$ \\
Max training steps         & $500{,}000$ \\
Gradient clipping (max-norm) & $1.0$ \\
Mixed precision            & bf16 (TF32 matmul) \\
Diffusion time $\epsilon_t$ & $10^{-3}$ \\
Chunking-loss weight       & $10^{-2}$ \\
Router-aux-loss weight     & $10^{-2}$ \\
\bottomrule
\end{tabular}
\end{table}

\subsection{Model Architecture}

All models share a qwen-style decoder backbone with RMSNorm ($\varepsilon{=}10^{-6}$), SiLU-gated MLPs, RoPE positional embeddings
($\theta{=}10^{6}$, max position $40{,}960$), grouped-query attention without
attention bias or dropout, and tied input/output embeddings (untied for the
$16$B MoE variant). 
The vocabulary size is $50{,}260$ for the WebText experiments and $151{,}936$ for the largest MoE configuration.
Initializer standard deviation is $0.02$. 
Dense MDLM uses FlashAttention-2; BDLM, DCDM and DCDM-MoE use FlexAttention to support block-/chunk-structured masking. 
BDLM applies block-wise diffusion with block size $B = 8$.
DCDM partitions the sequence into $K$ semantic chunks via the chunking attention layer of Section~\ref{sec:cha}, which uses $K$ learnable centroid
projections of subspace dimension $h$ together with the load-balancing
bias of Section~\ref{sec:load_balancing}. DCDM-MoE replaces every decoder
MLP with a Qwen-style sparse mixture of experts
(\texttt{decoder\_sparse\_step}{=}1) with top-$k$ routing (un-normalized)
plus a shared expert. The architectural hyperparameters of all
configurations used in this work are listed in
Table~\ref{tab:model-arch}.

\begin{table}[h]
\centering
\small
\setlength{\tabcolsep}{4pt}
\caption{
    Model architectures. $N_L$: number of decoder layers; 
    $d$: hidden size; 
    $d_{\text{ff}}$: dense FFN intermediate size; 
    $H/H_{kv}$: query/KV attention heads;
    $d_h$: per-head dimension;
    $B$: BDLM block size;
    $K$: number of DCDM chunks (Section~\ref{sec:cha});
    $h$: subspace dimension of the chunking clusers $\bm{\mu}_k \in \mathbb{R}^{d \times h}$;
    $E/k$: number of experts / experts per token;
    $d_{\text{moe}}$: per-expert intermediate size;
    $d_{\text{shared}}$: shared-expert intermediate size.
}
\label{tab:model-arch}
\begin{tabular}{lcccccccccccc}
\toprule
\textbf{Model} & $N_L$ & $d$ & $d_{\text{ff}}$ & $H$ & $H_{kv}$ & $d_h$
& $B$ & $K$ & $h$
& $E/k$ & $d_{\text{moe}}$ & $d_{\text{shared}}$ \\
\midrule
\multicolumn{13}{l}{\textit{Vanilla Diffusion MDLM}} \\
MDLM-0.1B          & 12 & 768  & 3072 & 12 & 12 & 128 & -- & -- & --  & -- & --   & --   \\
MDLM-0.5B          & 28 & 1024 & 3072 & 16 & 8  & 128 & -- & -- & --  & -- & --   & --   \\
MDLM-1.7B          & 28 & 2048 & 6144 & 16 & 8  & 128 & -- & -- & --  & -- & --   & --   \\
\midrule
\multicolumn{13}{l}{\textit{Block-diffusion BDLM}} \\
BDLM-0.1B          & 12 & 768  & 3072 & 12 & 12 & 128 & 8 & -- & --  & -- & --   & --   \\
BDLM-0.5B          & 28 & 1024 & 3072 & 16 & 8  & 128 & 8 & -- & --  & -- & --   & --   \\
BDLM-1.7B          & 28 & 2048 & 6144 & 16 & 8  & 128 & 8 & -- & --  & -- & --   & --   \\
\midrule
\multicolumn{13}{l}{\textit{Chunked-diffusion DCDM}} \\
DCDM-0.1B (K=8)    & 12 & 768  & 3072 & 12 & 12 & 128 & -- & 8  & 64  & -- & --   & --   \\
DCDM-0.5B (K=8)    & 28 & 1024 & 3072 & 16 & 8  & 128 & -- & 8  & 256 & -- & --   & --   \\
DCDM-0.5B (K=16)   & 28 & 1024 & 3072 & 16 & 8  & 128 & -- & 16 & 128 & -- & --   & --   \\
DCDM-0.5B (K=32)   & 28 & 1024 & 3072 & 16 & 8  & 128 & -- & 32 & 128 & -- & --   & --   \\
DCDM-0.5B (K=64)   & 28 & 1024 & 3072 & 16 & 8  & 128 & -- & 64 & 96  & -- & --   & --   \\
DCDM-1.7B (K=8)    & 28 & 2048 & 6144 & 16 & 8  & 128 & -- & 8  & 128 & -- & --   & --   \\
DCDM-1.7B (K=16)   & 28 & 2048 & 6144 & 16 & 8  & 128 & -- & 16 & 128 & -- & --   & --   \\
\midrule
\multicolumn{13}{l}{\textit{Sparse DCDM-MoE}} \\
DCDM-MoE-0.8B/A0.4B & 28 & 896  & 2688  & 16 & 8 & 128 & -- & 8  & 128 & 8/2   & 896  & 896  \\
DCDM-MoE-2.8B/A1.2B & 28 & 1792 & 5376  & 16 & 8 & 128 & -- & 12 & 192 & 8/2   & 1792 & 1792 \\
\bottomrule
\end{tabular}
\end{table}

\section{Additional Results}

\subsection{Point vs.\ Subspace Clusters}
\label{app:point-vs-subspace}

In Section~\ref{sec:cha} we argued that point-based clustering inside an attention layer is brittle in the high-dimensional embedding spaces of modern language models, and motivated the subspace parameterization of chunking attention as a remedy.
We verify this claim empirically with an ablation that isolates the dimensionality of the cluster representation within the chunking attention layer itself, leaving every other component of the model untouched.

\paragraph{Setup.}
We train two variants of DCDM at the 0.1B scale that are identical in every aspect except the per-cluster subspace dimension $h$. The \textbf{point} variant
uses $h = 1$: each centroid $\bm{\mu}_k$ collapses to a single direction in $\mathbb{R}^{d}$, and the associated ``subspace'' degenerates to the line it spans.
This is the minimal configuration of our parameterization and serves as a faithful
in-architecture analogue of point-based attention clustering. 
The \textbf{subspace} variant uses $h = 48$, the value adopted throughout the
rest of the paper. The training data, tokenizer, optimizer, learning-rate schedule, batch size, number of clusters $K$, load-balancing setup, and all other hyperparameters are held fixed across the two runs; the only difference is the trailing dimension of $\bm{\mu}_k$.

\begin{figure}[t]
    \centering
    \includegraphics[width=\linewidth]{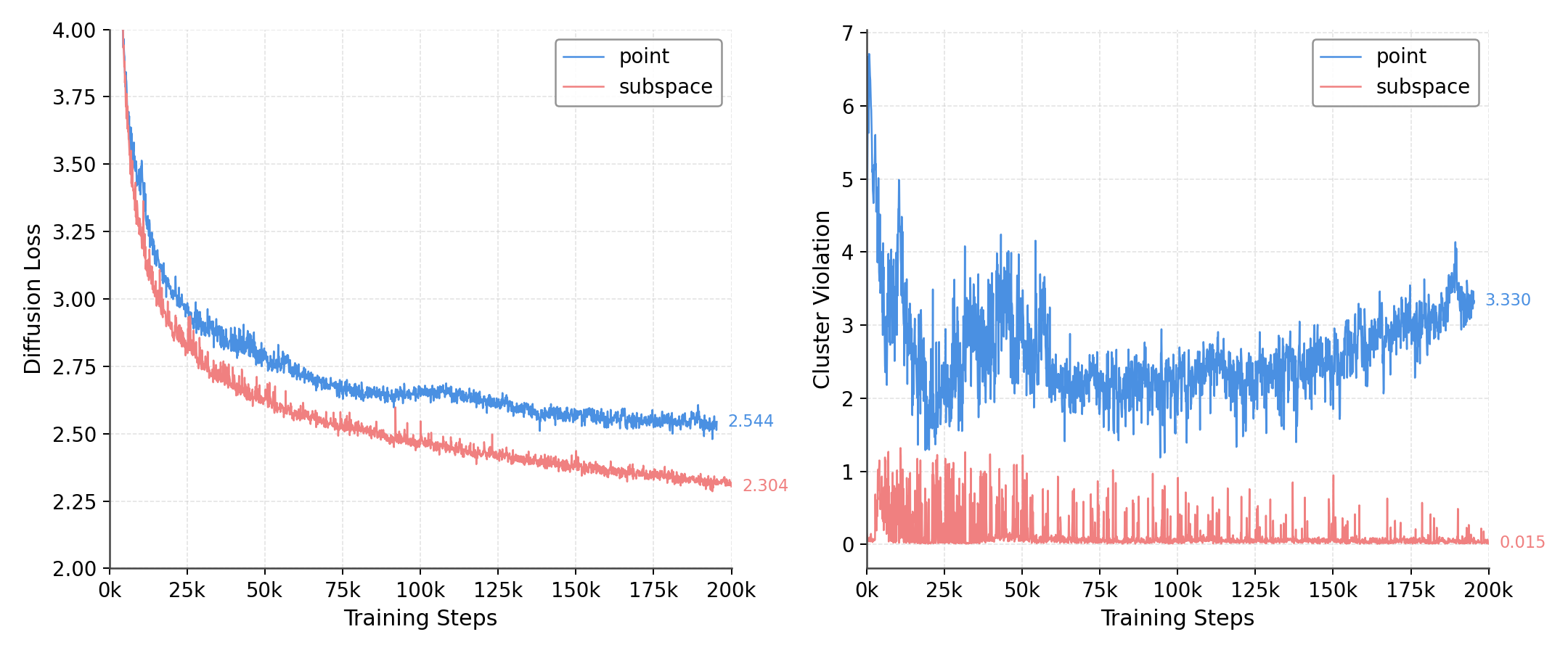}
    \caption{Point ($h = 1$) vs.\ subspace ($h = 48$) chunking attention
    at the 0.1B scale; all other training settings are identical.
    \textbf{Left:} Diffusion training loss. The subspace variant reaches
    a lower final loss ($2.304$ vs.\ $2.544$) and follows a smoother
    trajectory throughout training. \textbf{Right:} Cluster violation, a
    measure of deviation from uniform centroid usage (lower values
    indicate more uniform routing). The subspace variant drives violation
    to near zero within the first $\sim$10k steps and stays there;
    the point variant remains noisy throughout and rises sharply in the
    final 50k steps to $3.330$, the qualitative signature of progressive
    centroid collapse.}
    \label{fig:point-vs-subspace}
\end{figure}

\paragraph{Results.}
Figure~\ref{fig:point-vs-subspace} reports two diagnostics for both variants:
the diffusion training loss (left) and a cluster-violation metric tracking how far the centroid usage distribution deviates from uniform routing (right; values closer to zero indicate that the $K$ clusers receive comparable shares of tokens, while larger values indicate concentration of mass on a smaller subset).

The diffusion-loss panel matches the qualitative argument of Section~\ref{sec:cha}: the subspace variant reaches a substantially lower final loss ($2.304$ vs.\ $2.544$ at 200k steps), 
a gap that opens within the first 25k steps and persists throughout training, and its
trajectory is visibly less noisy.

The cluster-violation panel sharpens the picture from a quantitative gap into a qualitative failure mode. 
The subspace variant drives violation to near zero within the first $\sim$10k steps and remains essentially flat for the rest of training (final value $0.015$):
all $K$ clusers continue to receive a roughly uniform share of tokens, exactly as the auxiliary load-balancing loss intends. 
The point variant, in contrast, never manages to push violation below $\sim$2 during the entire run, and its violation begins to rise again past $\sim$125k steps, reaching $3.33$ by the end of training. 
The diffusion loss is near-stationary in this late regime while the routing
distribution is actively degrading: this is the qualitative signature of the centroid-collapse failure mode anticipated in Section~\ref{sec:cha},
in which a few clusers absorb most of the routed mass,
the remaining clusers are starved of gradient, and the auxiliary load-balancing loss is no longer sufficient to recover them.

Together, the two panels show that the difference between point-based and subspace clustering is not a matter of degree, a more capable variant settling at a slightly better minimum, but a qualitative one:
the subspace parameterization keeps routing well balanced throughout training, whereas the point parameterization actively degenerates even with the same load-balancing mechanisms in place.

\subsection{Zero-Shot Perplexity Evaluation}
\label{app:zero-shot-ppl}
 
As a complementary check on language-modeling quality, we evaluate the
diffusion models against a standard left-to-right autoregressive
transformer (\textbf{AR}) trained with the next-token objective. 
Following the convention of prior works~\citep{mdlm,bdlm}, all
four models are trained on OpenWebText~\cite{openwebtext} for 64B
tokens at the 0.1B parameter scale, and then evaluated, without any
further finetuning, on seven held-out corpora spanning encyclopedic,
news, scientific, and benchmark text: PTB~\cite{ptb}, WikiText~\cite{wiki}, LM1B~\cite{lm1b}, Lambada~\cite{lambada}, AG
News~\cite{news}, PubMed, and ArXiv~\cite{pub}.
 
\paragraph{Metric.}
We report token-level perplexity (lower is better). To ensure
comparability across models with different output parameterizations,
all perplexities are computed under a common tokenization and
evaluation context length, with the diffusion models evaluated using
the NELBO-based perplexity surrogate of~\citet{mdlm}.
 
\paragraph{Results.}
Table~\ref{tab:benchmark_ppl} reports the resulting zero-shot
perplexities. The autoregressive baseline retains an overall advantage
on perplexity, achieving the best score on PTB, WikiText, LM1B, and AG
News. This is consistent with the well-documented gap between AR
likelihoods and the NELBO-based perplexity surrogate used by all
diffusion models in the table~\citep{mdlm,bdlm}, so we focus the
comparison on the three diffusion models, where the metric is computed
under a common surrogate.
 
Among the diffusion models, DCDM improves over the unstructured baseline
MDLM on six of the seven corpora, with the largest reductions on PTB
($-3.14$) and PubMed ($-2.96$). The only corpus on which DCDM trails
MDLM is LM1B, where the gap is 1.16 points. Against the
positional-block baseline BDLM, DCDM is better on five of the seven
corpora, with WikiText and LM1B as the two exceptions. The gap is 0.63
points on WikiText but a more substantial 5.22 points on LM1B. On
Lambada, PubMed, and ArXiv, DCDM additionally achieves the lowest
perplexity among all four models, including AR. On balance, replacing
positional blocks with content-defined semantic blocks yields
consistent improvements in zero-shot perplexity within the
block-diffusion family, while preserving the broader gains of block
structure over unstructured masked diffusion.
 
\begin{table}[t]
    \centering
    \caption{
    Zero-shot validation perplexities ($\downarrow$) on OpenWebText.
    MDLM and BDLM are trained for 256B tokens, while DCDM is trained for 128B tokens.
    * indicates data borrowed from BDLM~\citep{bdlm}.
    }
    \label{tab:benchmark_ppl}
    \resizebox{\textwidth}{!}{
    \begin{tabular}{l|cccccccc}
    \toprule
     & Param & PTB & WikiText & LM1B & Lambada & AG News & PubMed & ArXiv \\
    \midrule
    \gray{AR*}
    & \gray{0.1B}    & \gray{81.07}   & \gray{25.32}    & \gray{51.14}
    & \gray{52.13}   & \gray{52.11}   & \gray{48.59}    & \gray{41.22}   \\
    \midrule
    MDLM*
    & 0.1B           & 90.96          & 33.22           & 64.94
    & 48.29          & 62.78          & 43.13           & 37.89          \\
    BDLM*
    & 0.1B           & 96.81          & \textbf{31.31}  & \textbf{60.88}
    & 50.03          & 61.67          & 42.52           & 39.20          \\
    DCDM
    & 0.1B           & \textbf{87.82} & 31.94           & 66.10
    & \textbf{46.43} & \textbf{59.85} & \textbf{40.17}  & \textbf{35.41} \\
    \bottomrule
    \end{tabular}
    }
\end{table}


\section{Limitation}
\label{app:limitation}
DCDM treats the number of semantic clusters $K$ as a fixed architectural
hyperparameter, shared across all sequences and all training stages.
This simplification keeps the chunking attention layer efficient and
admits stable optimization, but it also implies that the model uses the
same number of semantic blocks regardless of the length or complexity of
the input: sequences with little structural variety may end up
over-partitioned, while topically rich or unusually long sequences may be
under-partitioned. To avoid this tension we follow the standard practice
of related work in mixtures-of-experts and block diffusion language
models, fixing $K$ at training time and validating its choice via
ablation, and our experiments in Table~\ref{tab:k-ablation-full} confirm
that DCDM is robust to the precise value of $K$ within a reasonable
range, with all configurations tested improving over the positional-block
baseline at every scale we evaluated. Lifting this restriction by
learning the number of clusters per sequence, or by maintaining a
distribution over $K$ that can be marginalized at inference time, is a
natural extension and an interesting direction for future work.


\section{Impact}
\label{app:impact}

\paragraph{Ethical impacts.}
This work does not raise any direct ethical concerns. All experiments
are conducted on publicly available datasets, and the study involves
neither private data nor subjective human assessments at any stage.

\paragraph{Expected societal implications.}
The principal societal risk associated with DCDM is shared with any
capable generative language model: potential misuse for producing
misleading content, spam, or material that violates privacy or
intellectual-property norms. Mitigations developed for the broader class
of large language models, content provenance and watermarking,
deployment-time output filtering, and clear usage policies, apply
equally here, and we encourage practitioners building on this work to
adopt them.



\end{document}